\pdfoutput=1

\documentclass[11pt]{article}

\usepackage{acl}

\usepackage{times}
\usepackage{latexsym}
\usepackage{graphicx}
\usepackage{booktabs}
\usepackage{amsmath}
\usepackage{colortbl}
\usepackage[T1]{fontenc}

\usepackage[utf8]{inputenc}

\usepackage{microtype}

\usepackage{inconsolata}

\usepackage{multirow}
\title{Actor Identification in Discourse:  A Challenge for LLMs?}

\author{Ana Bari\'c$^*$$^\dagger$ \and Sebastian Pad\'o$^*$ \and Sean Papay$^*$ \\
    $^*$: IMS, University of Stuttgart, Stuttgart, Germany \\
    $^\dagger$: TakeLab, FER, University of Zagreb, Croatia \\
\texttt{\{ana.baric,sebastian.pado,sean.papay\}@ims.uni-stuttgart.de} \\
    }

\begin{document}
\maketitle

\begin{abstract}

The identification of political actors
who put forward claims in public debate is a crucial step in the
construction of \textit{discourse networks}, which are helpful to
analyze societal debates. Actor identification is, however, rather
challenging: Often, the locally mentioned speaker of a claim is only a pronoun (\textit{``He proposed that [claim]''}), so recovering the \textit{canonical} actor name requires discourse understanding.  
We compare a traditional pipeline of dedicated NLP components (similar to those applied to the related task
of coreference) with a LLM, which appears a good match for this generation task.
Evaluating on a corpus of German actors in newspaper reports, we find surprisingly that the LLM performs worse. Further analysis reveals that the LLM is very good at identifying the right reference, but struggles to generate the correct \textit{canonical form}.
This points to an underlying issue in LLMs with controlling generated output. Indeed, a hybrid model combining the LLM with a  classifier to normalize its output substantially outperforms both initial models.

\end{abstract}

\section{Introduction}

\begin{figure}[bt!]
    \centering
    \includegraphics[width=0.5\textwidth]{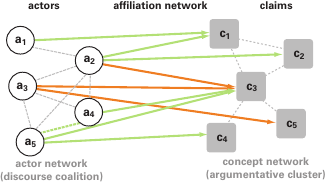}
    \caption{Discourse network with actors as circles and claims as squares (adapted from \citealp{pado-etal-2019-sides})}
    \label{fig:network-ex}
\end{figure}

Political decision-making in democracies is generally preceded by political debates taking place in parliamentary forums (committees, plenary debates) or different public spheres (e.g., newspapers, television, social media).
One way in which political scientists have analyzed such processes is to adopt the framework of political claims analysis \citep{koopmansPoliticalClaimsAnalysis1999}, identifying the \textit{claims} (i.e., calls for or against specific courses of action) and \textit{actors} involved in a given debate. Actors, claims, and the relations between them can then be represented as bipartite \textit{discourse networks}
\cite{Leifeld_Haunss_2012,leifeldDiscourseNetworkAnalysis2016}, such as shown in Figure~\ref{fig:network-ex}.
Such networks permit researchers to investigate debates on a fine-grained level, identifying, e.g., discourse coalitions, decision makers, or argumentative clusters.

While early work on discourse networks was based on manual analysis, widespread use of discourse networks requires quick, ideally automatic, methods to construct them from text.
This calls for NLP methods to (1) detect instances of claims, assign them to their categories (c$_i$ in Figure~\ref{fig:network-ex}),
and (2) identify actors for these claims in terms of some canonical representation (a$_i$), cf. \citet{pado-etal-2019-sides}.

At least for newswire, there are several NLP models for claim detection and categorization \citep{subramanian-etal-2018-hierarchical,pado-etal-2019-sides}.
In contrast, there is little work on actor identification. Arguably, this is because claims are easier to handle: Both detection and categorization are sentence-level classification tasks which can be modeled based on predominantly sentence-internal features.
In contrast, actor identification calls for a substantial amount of discourse understanding: models must \textit{locally} identify
an actor for the claim, but since these are often just a pronoun or a definite description (cf. Table\ \ref{tab:biden-ex}), they must \textit{globally} find 
a reasonable canonical representation for that actor.

\begin{table*}[tbh!]
    \centering
    \begin{tabular}{lll}
    \toprule
    & Local mention of actor & Canonical version \\
    \midrule
    1 & \textit{President Joe Biden} pleaded with Republicans \dots & Joe Biden \\
    2 & \textit{Biden} signaled a willingness to make significant changes \dots & Joe Biden \\
    3 & ``We can't let Putin win'', \textit{he} said. & Joe Biden \\
    4 & However, \textit{Senate Republicans} later on Wednesday blocked \dots & Senate Republicans \\
    5 & A \textit{U.S. official} said Washington had less than \$1B \dots & U.S. official \\
    \bottomrule
    \end{tabular}
    \caption{Actor mentions and their canonicalizations in newswire article (\url{https://shorturl.at/WZ159})}
    \label{tab:biden-ex}
\end{table*}

This paper asks whether this situation has improved with the emergence of prompt-based LLMs \citep{10.1145/3560815} and their promise for text-to-text generation, which appears to be a good match for the actor identification task. We contrast an LLM-based architecture with a traditionally trained pipeline of dedicated NLP components on a German dataset with actor-claim annotation \cite{blokker:_between}. We find that, surprisingly, the traditional architecture outperforms the LLM.
Our error analysis shows that the LLM often identifies the correct actor entity, but fails to generate the canonical actor name. We attribute this to the general difficulty in controling what exactly LLMs generate, a problem which has given rise to a substantial body of work \citep{zheng-etal-2023-invariant}. In line with this interpretation, we show that combining the LLM with the traditional model (for post-processing) achieves substantially better performance on the actor identification task than either model alone.

\section{Methods}

\subsection{Actor Identification: Task Definition}

Table \ref{tab:biden-ex} shows  mentions of actors making claims in a newswire article and the canonical actors they refer to, i.e.,\ input--output pairs for actor mapping.

One possible approach is to treat this task as entity linking \citep{10.3233/SW-222986}, typically realized as classification where the classes are the set of entities from a knowledge base (KB) such as Wikidata. While frequent actors (cf.\ lines 1--3) are mostly represented in such KBs, texts also introduce ad-hoc actors through plurals (line 4) or unspecific descriptions (line 5) which are generally not part of KBs. That rules out pure entity linking.

Instead, we formalize actor identification directly as \textit{canonical name string prediction}: Models are presented with a claim, along with its context within an article, and are tasked with predicting a string representing that actor.
For actors which commonly recur across claims, this string will be a canonical form of the actor's full name, while for singleton actors, this string will be the verbatim realization of an actor mention from the article.

While this formalization seems to ignore much of the structure of the task (after all, actor names are not fundamentally arbitrary strings),
it has the benefit of allowing fair comparisons between vastly different model architectures: Text generation models can produce short strings directly, and other modeling approaches can take advantage of task structure internally, while still outputting a string. For example, we could approach the task with a coreference model, extended with a component which chooses the most canonical realization in each coreference chain from among the mentions.\footnote{We do not evaluate a coreference model since full coreference is known to be a very hard task \citep[see, e.g.,\ ][]{peng-etal-2015-solving} and actor identification only requires solving a subpart.}

\subsection{A Traditional Pipeline Architecture}

The first method we apply to this task is a pipeline of two
``traditional'' NLP approaches: an entity extractor for actor mentions, and a classifier for associating mentions with canonical actor names.

Our mention extractor is a CRF-based sequence labeler.
As input, we provide full articles in which the target claim has been marked
and encode the input with a pretrained XLM-RoBERTa encoder \citep{conneau-etal-2020-unsupervised}, which we fine-tune during training. The CRF's task is to extract mentions of the actor for the marked claim.
As each claim must have at least one actor mention, we constrain \citep{papay2022constraining} our CRF to always predict at least one actor mention.
In order to map actor mentions to canonical forms, we employ a simple neural classifier based on the same XLM-RoBERTa encoder as above. As classes, we use the set of all canonical actor names which occur at least twice in the training partition of our data (see Section~\ref{sec:data}), along with a special `verbatim' class for the remaining cases. In these cases, the string output we predict 
is the exact text of the actor mention.

\subsection{An LLM-Based Architecture}
\label{sec:an-llm-based}

In our LLM-based approach, we treat actor identification as an end-to-end task by combining the subtasks of actor detection and mapping within the prompt to directly predict the canonicalized actor. Due to the limited availability of language-specific LLMs, we opted to experiment with the Llama~2 language model \cite{touvron2023llama} for both base- and instruction model options in all available size variants. This model family could be used on German, despite being predominantly trained on English corpora, because of the cross-lingual transferability that is shown to occur in such multilingual LLMs \cite{choenni2023languages}.

We assess this task in zero- and few-shot settings, employing current
best practices for robust prompt construction. These include: (1)
using different instruction paraphrases for prompt templates, given
the fact that 'canonical name' is not a very established concept
(cf. Appendix \ref{sec:appendix}); (2) selecting exemplars
semantically similar to the input \cite{margatina2023active}; and (3)
varying exemplar quantity and order within the prompt
\cite{lu-etal-2022-fantastically}. We construct the prompts by
combining the English task description as prompt instruction with the
preprocessed article in German (again, cf. Appendix \ref{sec:appendix}). Due
to the context length limitation, we preprocess articles by extracting
the target claim, marked with special tags, with its surrounding
context at the sentence level. We use greedy decoding.

In these trials, zero-shot Llama-2-70b-chat  outperforms all few-shot settings. We choose this setting for the rest of the paper.

\section{Experimental Setup}

\subsection{Data}
\label{sec:data}

As gold standard for our studies we use DEbateNet \cite{blokker:_between}, a German large corpus resource for the analysis of the  domestic debate on migration in Germany in 2015. After domain experts from political science developed a codebook for the policy domain, roughly 700 newspaper articles from the German left-wing quality newspaper “taz – die tageszeitung” with a total of over 550,000 tokens were annotated for actors, claims, and their relations.
For each article, all claims are marked and labeled,
and each claim is associated with a canonical actor (our gold standard), yielding a collection of about 1,800 actor-attributed claims.  Most claims are also associated with a named entity mention from the vicinity of the claim,  though this may not be the nearest mention, cf. Table\ \ref{tab:biden-ex}.
We use the established DEbateNet train--dev--test split, with 1383 claims in train, 220 in dev, and 207 in test.

\subsection{Evaluation}
Both models are evaluated and compared via  $F_1$-score. In order to gain a more detailed understanding, we use three evaluation settings:

In the strictest \textit{exact-match} setting, predictions are counted as correct only if they exactly match the gold-standard actor string. This setting can be performed automatically.

In our \textit{correct-up-to-formatting} setting, predictions are counted as correct if they match the gold standard string modulo text formatting differences (e.g.\ whitespace differences, capitalization, punctuation). This setting tells us how often a model is ``almost right'' but receives no credit in the strict setting. We carry out this evaluation manually.

Finally, our \textit{correct-up-to-canonicalization} setting counts predictions as correct if they predict the correct entity, even if a different referring expression is generated. For example, ``the chancellor'' or ``Merkel'' would be considered correct predictions for the gold-standard actor ``Angela Merkel.'' As with before, this evaluation is performed manually.

\section{Results and Analysis}

\begin{table}
\setlength{\tabcolsep}{3.5pt}
    \begin{tabular}{llrrr}
    \toprule
    & Evaluation & Pr & Re & $F_1$ \\
    \midrule
    \multirow{3}{*}{LLM} & exact match & 42.66 & 43.46 & 43.06 \\
    & up to formatting & 43.56 & 44.39 & 43.98\\
    & up to canonic. & 62.39 & 63.55 & 62.96 \\
    \midrule
    \multirow{3}{*}{\shortstack{dedicated\\pipeline}} & exact match & 48.66 & 59.35 & 53.47\\
    & up to formatting & 48.66 & 59.35 & 53.47\\
    & up to canonic. & 54.79 & 66.82 & 60.21 \\
    \bottomrule
    \end{tabular}
    \caption{Results for the LLM and traditional pipeline models in the different evaluation settings
    \label{tab:results}
    }
\end{table}

\paragraph{Main results.} Table~\ref{tab:results} summarizes the performance of our two models under our three evaluation settings. We first consider our strictest setting, exact match. We find results in the range of 40--50 points $F_1$ score, in line with the assumption that actor mapping is a difficult task. Both models have somewhat higher recall than precision, and the dedicated pipeline outperforms the LLM by 10 point $F_1$ score. This is somewhat surprising, given LLMs' well-known capabilities in instruction-following text generation
\citep{NEURIPS2020_1457c0d6, webson2022promptbased, zhou2023instructionfollowing}.


We form two non-mutually exclusive hypotheses for this performance gap: either that the traditional model, through its supervised training, came to be more competent at predicting the \textit{correct} political actor, or, through virtue of its inductive biases, it came to better and predicting the \textit{exact} canonical name.
We examine these hypotheses by evaluating the model with the other two settings. We also carry out a qualitative analysis of errors made by the LLM-based model (see Table~\ref{tab:eg}).

One simple factor that would lead an essentially correct LLM to be inexact is formatting errors in its output -- either mismatched spacing, punctuation, or capitalization, or natural language responses that could not be correctly post-processed. Such effects should 
show up as a difference between the `exact match' and the `up to formatting' setting. 
However, the numbers (43.06 $F_1$ vs. 43.98 $F_1$) show that these types of error account for less than one percentage point.
Our qualitative error analysis (Table~\ref{tab:eg}, top part) finds (few) cases of formatting errors, which often co-occur with other problems (unexpected LLM responses, gold standard errors). We conclude that such errors have a relatively minor effect on performance.

\begin{table}[tb]
    \setlength{\tabcolsep}{5pt}
    \begin{tabular}{lp{3.5cm}p{2.2cm}}
    \toprule
    \textbf{\shortstack{Error\\Type}} & \textbf{Model output} & \textbf{Ground Truth} \\
    \midrule
    \multirow{2}{*}[-1em]{\rotatebox{90}{Format}} 
    &       \cellcolor{lightgray!20}  Bayern The claim is & \cellcolor{lightgray!20}Bayern \textit{(Bavaria)} \\
    &EU-Kommission \par \textit{(EU commission)} & EU-Kommision  [\textit{sic}] \\ 
    \addlinespace
    \midrule
    \addlinespace
    \multirow{5}{*}[-3.5em]{\rotatebox{90}{Canonicalization}}
    &        \cellcolor{lightgray!20} Bundesinnenminister \textit{(federal minister of the interior)} Thomas de Maizi\`ere & \cellcolor{lightgray!20}  Thomas de Maizi\`ere \\
    &Kommissions-\par pr\"asident (\textit{commission president}) Jean-Claude Juncker & Jean-Claude Juncker \\
    &         \cellcolor{lightgray!20} 
Zimmermann & \cellcolor{lightgray!20} Klaus F.\ Zimmermann\\
    &Merkel & Angela Merkel \\
    \addlinespace
    \midrule
    \addlinespace
    \multirow{2}{*}[-.4em]{\rotatebox{90}{\shortstack{Wrong\\Actor}}}
     & \cellcolor{lightgray!20}  EU-Kommission \par \textit{(EU commission)} &    \cellcolor{lightgray!20}  Jean-Claude Juncker
    \\
    & Germany & Thomas Bauer \\
    \bottomrule
    \end{tabular}
    \caption{
    \label{tab:eg} Some illustrative examples of the errors exhibited by the LLM-based actor identification model: German outputs with English translations}
\end{table}

The reliance of our exact evaluation metric on gold-standard canonical
forms provides another opportunity for a largely correct model to show
low performance due to an inability to pick the exact canonical form
required. This factor should come to the fore when we compare exact
match results to the 'up-to-canonicalization' setting. Indeed, for
this setting, both models show a substantial increase in performance
-- which implies that canonicalization represents a large part of the
difficulty for this task. Interestingly, the LLM shows a much larger
improvement, ultimately outperforming the traditional pipeline by
about 2.5 points $F_1$.  Our qualitative error analysis in
Table~\ref{tab:eg} (center part) indicates that our LLM predictions
have a hard time hitting the right level of verbosity: they are either
too verbose, spuriously including government positions (e.g.\
[\textit{Interior Minister}] \textit{Thomas de Maizi\`ere}), or not
verbose enough, omitting first names (e.g.\ [\textit{Angela}]
\textit{Merkel}).

We take this as evidence that our LLM-based model is adept at
selecting the correct actor, but struggles to select the canonical
form.  This is somewhat to be expected, as our LLM-based model has
neither a training signal nor a strong inductive bias to prefer any
particular canonical form.  However, as mentioned in
Section~\ref{sec:an-llm-based}, preliminary experiments with a
few-shot setting where we included canonical forms in prompts showed
no improvements over our proposed model. We believe that this
indicates that the task of predicting 'canonical names' remains a
non-straightforward task for LLMs even in the presence of training
data.

Finally, responses which bungled the reference completely (Table~\ref{tab:eg}, bottom part) sometimes tended to be plausible, e.g. metonymyic, mistakes, such as predicting the EU commission instead of Jean-Claude Juncker, its president.

\paragraph{Hybrid model.} The observations on the errors motivate a follow-up experiment with a hybrid approach combining both our traditional and LLM-based models.
This hybrid is structurally similar to our traditional model, but it is provided the LLM's prediction in addition to its other inputs. In this way, the LLM can decide which actor made the claim, while the traditional pipeline can be responsible for predicting that actor in a canonical form. Table \ref{tab:hybrid} shows that this 
approach has similar properties to the individual models
(no effect of formatting, but a large effect of canonicalization) but that it represents, crucially, a substantial improvement in terms of quality: In the strictest setting (exact match), it achieves an $F_1$ score of 59 points (previous best: 53 F$_1$), and in the laxest setting it obtains 70 points $F_1$ (previous best: 63 F$_1$).

\begin{table}[t!b]
    \begin{tabular}{lrrr}
    \toprule
    Evaluation & Pr & Re & $F_1$ \\
    \midrule
    exact match & 54.33 & 64.49 & 58.97 \\ 
    up to formatting & 54.33 & 64.49 & 58.97 \\
    up to canonic. & 64.96 & 76.39 & 70.21 \\
    \bottomrule
    \end{tabular}
    \caption{
    \label{tab:hybrid}Results for the hybrid model in the different evaluation settings}
\end{table}

\section{Conclusion}

In this work, we investigate alternative approaches to tackling the discourse-level actor identification task, comparing LLM prompting with a conventional NLP pipeline. 
We find that our LLM better recognize the appropriate actor entities compared to the traditional pipeline, but has a harder time controlling the exact output. This problem cannot be solved easily with tuning, as the failure of our few-shot setup shows, which is also in line with recent studies on the controllability of LLM output
\citep{reif-etal-2022-recipe,sun-etal-2023-evaluating}. Our solution is a hybrid model which integrates the LLM-generated output as a cue in the pipeline approach, resulting in a clear improvement over the individual models.

The current study is limited in several respects: It only considers one LLM, one corpus, and one evaluation. 
In the future, we also plan to carry out an extrinsic evaluation 
of our actor identifier on generating full discourse networks. In terms of future directions, we believe that
actor identification is a task which could plausibly profit 
from retrieval-augmented generation (RAG) proposed by \citet{lewis2021retrievalaugmented} which would give the
LLM access to information beyond the current discourse.

\section*{Acknowledgements}

We acknowledge funding by the Deutsche Forschungsgemeinschaft (DFG) for the
project MARDY 2 (375875969) within the priority program RATIO.

\bibliography{custom}
\onecolumn
\appendix
\section{Prompt Templates}
\label{sec:appendix}

\begin{table*}[h]
\begin{tabular}{p{0.5cm}p{13.2cm}}
\toprule
\textbf{\#} &
  \textbf{Instruction templates} \\ \midrule
1 &
  \textit{"Extract only the entity that made the claim in the article. The claim is surrounded with \textless{}claim\textgreater and \textless{}\textbackslash{}claim\textgreater tags. Output only the entity without any additional explanation. Article: {[}ARTICLE{]}"} \\
  \addlinespace
2 &
  \textit{"Extract and standardize only the entity that made the marked claim in the article. The claim is surrounded with \textless{}claim\textgreater and \textless{}\textbackslash{}claim\textgreater tags. Output only the standardized entity without any additional explanation. Article: {[}ARTICLE{]}"} \\
  \addlinespace
3 &
  \textit{"Retrieve the party or parties responsible for the statement in the given article, contained within \textless{}claim\textgreater and \textless{}\textbackslash{}claim\textgreater 
 tags. Output only the entity without further elaboration. Article:{[}ARTICLE{]}"} \\
  \addlinespace
4 &
  \textit{"Identify and output the entity or entities that made the claim within the specified article, enclosed by \textless{}claim\textgreater and \textless{}\textbackslash{}claim\textgreater tags. Do not include any supplementary information. Article: {[}ARTICLE{]}"} \\ \bottomrule
\end{tabular}
\label{tab:prompt_templates}

\caption{Prompt template instruction paraphrases used for robustness check for zero- and few-shot setting.}
\end{table*}


\end{document}